\documentclass[5p,twocolumn]{elsarticle}

\usepackage{lipsum}
\makeatletter
\def\ps@pprintTitle{%
 \let\@oddhead\@empty
 \let\@evenhead\@empty
 \def\@oddfoot{}%
 \let\@evenfoot\@oddfoot}
\makeatother

\usepackage{amsmath}
\usepackage{amssymb}

\newtheorem{definition}{Definition}

\newcommand{\x}{\boldsymbol{x}}
\newcommand{\w}{\boldsymbol{w}}
\newcommand{\z}{\boldsymbol{z}}

\begin{document}

\begin{frontmatter}

\title{Methods for Interpreting and Understanding Deep Neural Networks}

\author[tu]{Gr\'egoire Montavon\corref{cor1}}
\ead{gregoire.montavon@tu-berlin.de}

\author[hhi]{Wojciech Samek\corref{cor1}}
\ead{wojciech.samek@hhi.fraunhofer.de}

\author[tu,kor,mpi]{Klaus-Robert M\"uller\corref{cor1}}
\ead{klaus-robert.mueller@tu-berlin.de}

\address[tu]{
Department of Electrical Engineering \& Computer Science, Technische Universit\"at Berlin, Marchstr. 23, Berlin 10587, Germany
}
\address[hhi]{
Department of Video Coding \& Analytics, Fraunhofer Heinrich Hertz Institute, Einsteinufer 37, Berlin 10587, Germany
}
\address[kor]{
Department of Brain \& Cognitive Engineering, Korea University, Anam-dong 5ga, Seongbuk-gu, Seoul 136-713, South Korea
}
\address[mpi]{
Max Planck Institute for Informatics, Stuhlsatzenhausweg, Saarbr\"ucken 66123, Germany
}

\cortext[cor1]{Corresponding authors}

\begin{abstract}
This paper provides an entry point to the problem of interpreting a deep neural network model and explaining its predictions. It is based on a tutorial given at ICASSP 2017. It introduces some recently proposed techniques of interpretation, along with theory, tricks and recommendations, to make most efficient use of these techniques on real data. It also discusses a number of practical applications.
\end{abstract}


\end{frontmatter}

\section{Introduction}

Machine learning techniques such as deep neural networks have become an indispensable tool for a wide range of applications such as image classification, speech recognition, or natural language processing. These techniques have achieved extremely high predictive accuracy, in many cases, on par with human performance.

In practice, it is also essential to verify for a given task, that the high measured accuracy results from the use of a proper problem representation, and not from the exploitation of artifacts in the data \citep{Leek2010, 10.1371/journal.pone.0100335,DBLP:conf/cvpr/LapuschkinBMMS16}. Techniques for interpreting and understanding what the model has learned have therefore become a key ingredient of a robust validation procedure \citep{Taylor:2005:MPV:1076653,DBLP:journals/jmlr/BaehrensSHKHM10,10.1371/journal.pone.0130140}. Interpretability is especially important in applications such as medicine or self-driving cars, where the reliance of the model on the correct features must be guaranteed \citep{DBLP:conf/kdd/CaruanaLGKSE15, DBLP:journals/corr/BojarskiYCCFJM17}.

It has been a common belief, that simple models provide higher interpretability than complex ones. Linear models or basic decision trees still dominate in many applications for this reason. This belief is however challenged by recent work, in which carefully designed interpretation techniques have shed light on some of the most complex and deepest machine learning models \citep{DBLP:journals/corr/SimonyanVZ13, DBLP:conf/eccv/ZeilerF14, 10.1371/journal.pone.0130140, DBLP:conf/nips/NguyenDYBC16, DBLP:conf/kdd/Ribeiro0G16}.

Techniques of interpretation are also becoming increasingly popular as a tool for exploration and analysis in the sciences. In combination with deep nonlinear machine learning models, they have been able to extract new insights from complex physical, chemical, or biological systems \citep{Hansen2011, Hansen2015, Sturm2016, Schutt2017, 10.1371/journal.pone.0174392}.

This tutorial gives an overview of techniques for interpreting complex machine learning models, with a focus on deep neural networks (DNN). It starts by discussing the problem of interpreting modeled concepts (e.g.\ predicted classes), and then moves to the problem of explaining individual decisions made by the model. The tutorial abstracts from the exact neural network structure and domain of application, in order to focus on the more conceptual aspects that underlie the success of these techniques in practical applications.

\section{Preliminaries}

Techniques of interpretation have been applied to a wide range of practical problems, and various meanings have been attached to terms such as ``understanding'', ``interpreting'', or ``explaining''. See \citep{DBLP:journals/corr/Lipton16a} for a discussion. As a first step, it can be useful to clarify the meaning we associate to these words in this tutorial, as well as the type of techniques that are covered.

We will focus in this tutorial on {\em post-hoc interpretability}, i.e.\ a trained model is given and our goal is to understand what the model predicts (e.g.\ categories) in terms what is readily interpretable (e.g.\ the input variables) \citep{10.1371/journal.pone.0130140,DBLP:conf/kdd/Ribeiro0G16}. Post-hoc interpretability should be contrasted to incorporating interpretability directly into the structure of the model, as done, for example, in \citep{DBLP:conf/aaai/PoulinESLGWFPMA06,DBLP:conf/kdd/CaruanaLGKSE15}.

Also, when using the word ``understanding'', we refer to a {\em functional understanding} of the model, in contrast to a lower-level mechanistic or algorithmic understanding of it. That is, we seek to characterize the model's black-box behavior, without however trying to elucidate its inner workings or shed light on its internal representations.

Throughout this tutorial, we will also make a distinction between {\em interpretation} and {\em explanation}, by defining these words as follows.

\begin{definition}
An interpretation is the mapping of an abstract concept (e.g.\ a predicted class) into a domain that the human can make sense of.
\end{definition}
Examples of domains that are interpretable are images (arrays of pixels), or texts (sequences of words). A human can look at them and read them respectively. Examples of domains that are \emph{not} interpretable are abstract vector spaces (e.g. word embeddings \citep{DBLP:conf/nips/MikolovSCCD13}), or domains composed of undocumented input features (e.g.\ sequences with unknown words or symbols).

\begin{definition}
An explanation is the collection of features of the interpretable domain, that have contributed for a given example to produce a decision (e.g.\ classification or regression).
\end{definition}
An explanation can be, for example, a heatmap highlighting which pixels of the input image most strongly support the classification decision \citep{DBLP:journals/corr/SimonyanVZ13,DBLP:conf/cidm/LandeckerTBMKB13,10.1371/journal.pone.0130140}. The explanation can be coarse-grained to highlight e.g.\ which regions of the image support the decision. It can also be computed at a finer grain, e.g.\ to include pixels and their color components in the explanation. In natural language processing, explanations can take the form of highlighted text \citep{DBLP:conf/naacl/LiCHJ16,DBLP:journals/corr/ArrasHMMS16a}.

\section{Interpreting a DNN Model}

This section focuses on the problem of interpreting a concept learned by a deep neural network (DNN). A DNN is a collection of neurons organized in a sequence of multiple layers, where neurons receive as input the neuron activations from the previous layer, and perform a simple computation (e.g.\ a weighted sum of the input followed by a nonlinear activation). The neurons of the network jointly implement a complex nonlinear mapping from the input to the output. This mapping is learned from the data by adapting the weights of each neuron using a technique called error backpropagation~\citep{Rumelhart1986}.

The learned concept that must be interpreted is usually represented by a neuron in the top layer. Top-layer neurons are abstract (i.e.\ we cannot look at them), on the other hand, the input domain of the DNN (e.g.\ image or text) is usually interpretable. We describe below how to build a {\em prototype} in the input domain that is interpretable and representative of the abstract learned concept. Building the prototype can be formulated within the activation maximization framework.

\subsection{Activation Maximization (AM)}
\label{section:interpretation-l2}

Activation maximization is an analysis framework that searches for an input pattern that produces a maximum model response for a quantity of interest \citep{DBLP:journals/neco/BerkesW06,Erhan2009,DBLP:journals/corr/SimonyanVZ13}.

Consider a DNN classifier mapping data points $\x$ to a set of classes $(\omega_c)_c$. The output neurons encode the modeled class probabilities $p(\omega_c|\x)$. A prototype $\x^\star$ representative of the class $\omega_c$ can be found by optimizing:
$$
\max_{\x} ~ \log p(\omega_c|\x) - \lambda \|\x\|^2.
$$
The class probabilities modeled by the DNN are functions with a gradient \citep{Bishop:1995:NNP:525960}. This allows for optimizing the objective by gradient ascent. The rightmost term of the objective is an $\ell_2$-norm regularizer that implements a preference for inputs that are close to the origin. When applied to image classification, prototypes thus take the form of mostly gray images, with only a few edge and color patterns at strategic locations \citep{DBLP:journals/corr/SimonyanVZ13}. These prototypes, although producing strong class response, look in many cases unnatural.

\subsection{Improving AM with an Expert}
\label{section:interpretation-expert}

In order to focus on more probable regions of the input space, the $\ell_2$-norm regularizer can be replaced by a data density model $p(\x)$ called ``expert'', leading to the new optimization problem: 
$$
\max_{\x} ~ \log p(\omega_c|\x) + \log p(\x).
$$
Here, the prototype is encouraged to simultaneously produce strong class response and to resemble the data. By application of the Bayes' rule, the newly defined objective can be identified, up to modeling errors and a constant term, as the class-conditioned data density $p(\x|\omega_c)$. The learned prototype thus corresponds to the most likely input $\x$ for class $\omega_c$. A possible choice for the expert is the Gaussian RBM \citep{DBLP:series/lncs/Hinton12}. Its probability function can be written as:
$$
\log p(\x) = {\textstyle \sum_j} f_j(\x) - {\textstyle \frac12} \x^\top \Sigma^{-1} \x + \text{cst.}
$$
where $f_j(\x) = \log(1+\exp(\w_j^\top \x + b_j))$ are factors with parameters learned from the data. When interpreting concepts such as natural images classes, more complex density models such as convolutional RBM/DBMs \citep{DBLP:conf/icml/LeeGRN09}, or pixel-RNNs \citep{DBLP:conf/icml/OordKK16} are needed.

\begin{figure}[h!]\centering
\includegraphics[width=0.97\linewidth]{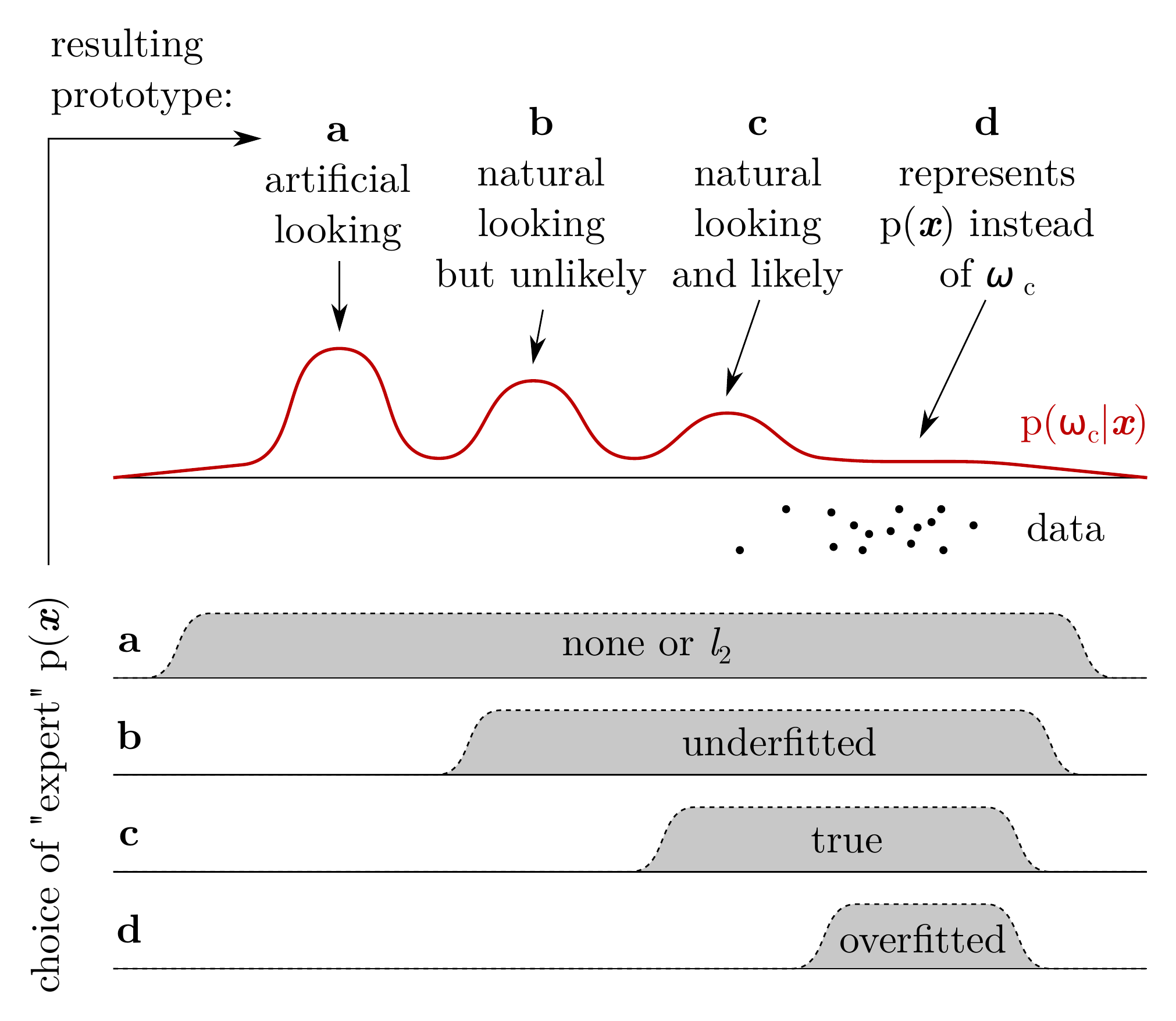}
\caption{Cartoon illustrating how the expert $p(\x)$ affects the prototype $\x^\star$ found by AM. The horizontal axis represents the input space, and the vertical axis represents the probability.}
\label{figure:expert}
\end{figure}

In practice, the choice of the expert $p(\x)$ plays an important role. The relation between the expert and the resulting prototype is given qualitatively in Figure~\ref{figure:expert}, where four cases (a--d) are identified. On one extreme, the expert is coarse, or simply absent, in which case, the optimization problem reduces to the maximization of the class probability function $p(\omega_c|\x)$. On the other extreme, the expert is overfitted on some data distribution, and thus, the optimization problem becomes essentially the maximization of the expert $p(\x)$ itself.
When using AM for the purpose of model validation, an overfitted expert (case d) must be especially avoided, as the latter could hide interesting failure modes of the model $p(\omega_c|\x)$. A slightly underfitted expert (case b), e.g.\ that simply favors images with natural colors, can already be sufficient. On the other hand, when using AM to gain knowledge on a correctly predicted concept $\omega_c$, the focus should be to prevent underfitting. Indeed, an underfitted expert would expose optima of $p(\omega_c|\x)$ potentially distant from the data, and therefore, the prototype $\x^\star$ would not be truly representative of $\omega_c$.

\subsection{Performing AM in Code Space}
\label{section:interpretation-code}

In certain applications, data density models $p(\x)$ can be hard to learn up to high accuracy, or very complex such that maximizing them becomes difficult. An alternative class of unsupervised models are generative models. They do not provide the density function directly, but are able to sample from it, usually via the following two steps:
\begin{enumerate}
\item Sample from a simple distribution $q(\z) \sim \mathcal{N}(0,I)$ defined in some abstract code space $\mathcal{Z}$.
\item Apply to the sample a decoding function $g:\mathcal{Z} \to \mathcal{X}$, that maps it back to the original input domain.
\end{enumerate}
One such model is the generative adversarial network \citep{DBLP:conf/nips/GoodfellowPMXWOCB14}. It learns a decoding function $g$ such that the generated data distribution is as hard as possible to discriminate from the true data distribution. The decoding function $g$ is learned in competition with a discriminant between the generated and the true distributions. The decoding function and the discriminant are typically chosen to be multilayer neural networks.

\citet{DBLP:conf/nips/NguyenDYBC16} proposed to build a prototype for $\omega_c$ by incorporating such generative model in the activation maximization framework. The optimization problem is redefined as:
$$
\max_{\z \in \mathcal{Z}}~ \log p(\omega_c \,|\, g(\z)) - \lambda \|\z\|^2,
$$
where the first term is a composition of the newly introduced decoder and the original classifier, and where the second term is an $\ell_2$-norm regularizer in the code space. Once a solution $\z^\star$ to the optimization problem is found, the prototype for $\omega_c$ is obtained by decoding the solution, that is, $\x^\star = g(\z^\star)$. In Section \ref{section:interpretation-l2}, the $\ell_2$-norm regularizer in the input space was understood in the context of image data as favoring gray-looking images. The effect of the $\ell_2$-norm regularizer in the code space can instead be understood as encouraging codes that have high probability. Note however, that high probability codes do not necessarily map to high density regions of the input space.

\begin{figure}[h!]
\centering
\includegraphics[width=0.99\linewidth]{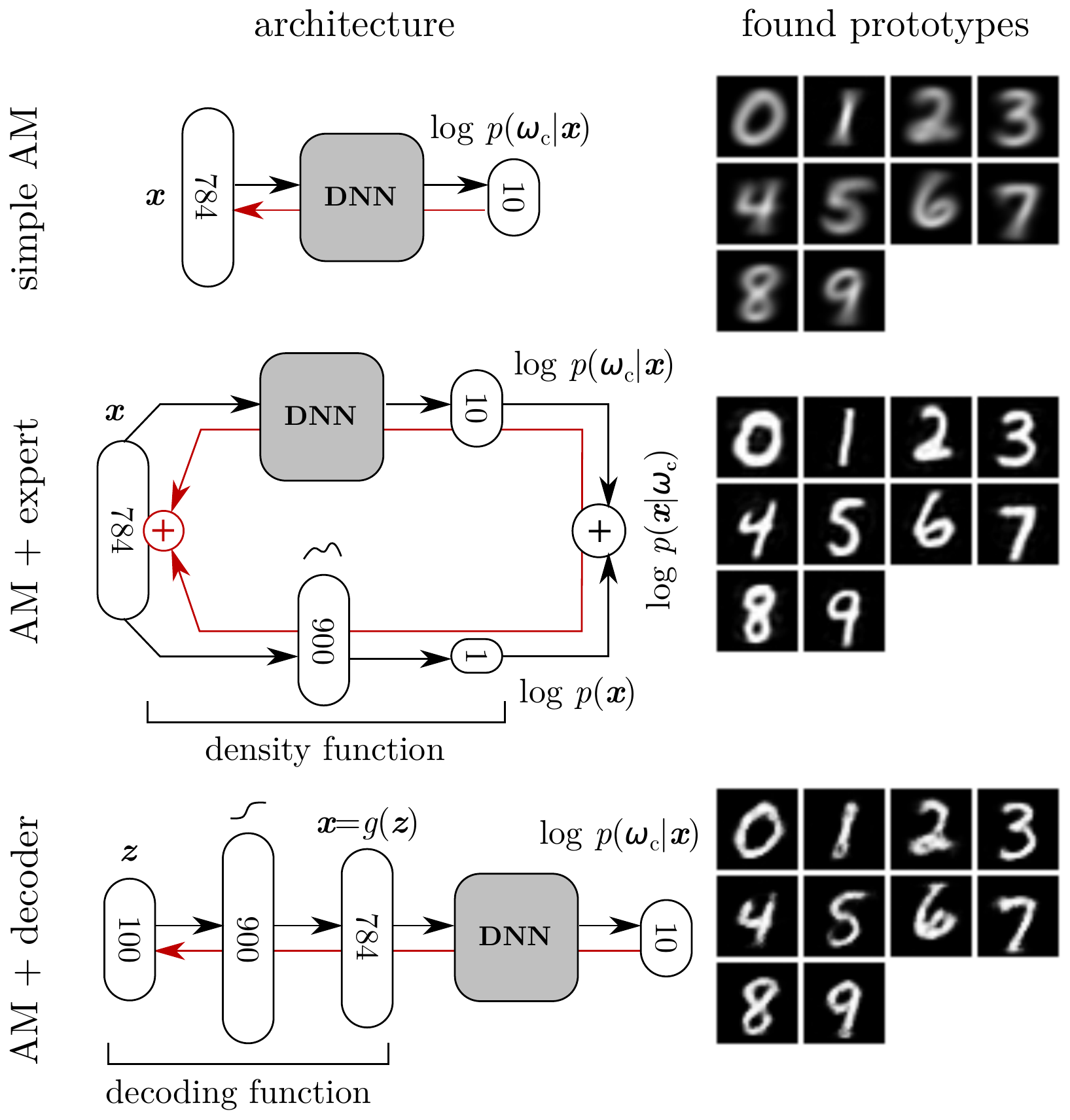}
\caption{Architectures supporting AM procedures and found prototypes. Black arrows indicate the forward path and red arrows indicate the reverse path for gradient computation.}
\label{figure:architectures}
\end{figure}

To illustrate the qualitative differences between the methods of Sections \ref{section:interpretation-l2}--\ref{section:interpretation-code}, we consider the problem of interpreting MNIST classes as modeled by a three-layer DNN. We consider for this task (1) a simple $\ell_2$-norm regularizer $\lambda \|\x-\bar{\x}\|^2$ where $\bar{\x}$ denotes the data mean for $\omega_c$, (2) a Gaussian RBM expert $p(\x)$, and (3) a generative model with a two-layer decoding function, and the $\ell_2$-norm regularizer $\lambda \|\z-\bar{\z}\|^2$ where $\bar{\z}$ denotes the code mean for $\omega_c$. Corresponding architectures and found prototypes are shown in Figure~\ref{figure:architectures}.
Each prototype is classified with full certainty by the DNN. However, only with an expert or a decoding function, the prototypes become sharp and realistic-looking.

\subsection{From Global to Local Analysis}

When considering complex machine learning problems, probability functions $p(\omega_c|\x)$ and $p(\x)$ might be multimodal or strongly elongated, so that no single prototype $\x^\star$ fully represents the modeled concept $\omega_c$. The issue of multimodality is raised by \citet{DBLP:journals/corr/NguyenYC16}, who demonstrate in the context of image classification, the benefit of interpreting a class $\omega_c$ using multiple local prototypes instead of a single global one.

Producing an exhaustive description of the modeled concept $\omega_c$ is however not always necessary. One might instead focus on a particular region of the input space. For example, biomedical data is best analyzed conditioned on a certain development stage of a medical condition, or in relation to a given subject or organ.

An expedient way of introducing locality into the analysis is to add a localization term $\eta \cdot \|\x - \x_0\|^2$ to the AM objective, where $\x_0$ is a reference point. The parameter $\eta$ controls the amount of localization. As this parameter increases, the question ``{\em what is a good prototype of $\omega_c$?}'' becomes however insubstantial, as the prototype $\x^\star$ converges to $\x_0$ and thus looses its information content.

Instead, when trying to interpret the concept $\omega_c$ locally, a more relevant question to ask is ``{\em what features of $\x$ make it representative of the concept $\omega_c$?}''. This question gives rise to a second type of analysis, that will be the focus of the rest of this tutorial.

\section{Explaining DNN Decisions}
\label{section:explanation}

\begin{figure}[h!]
\centering
\includegraphics[width=0.97\linewidth]{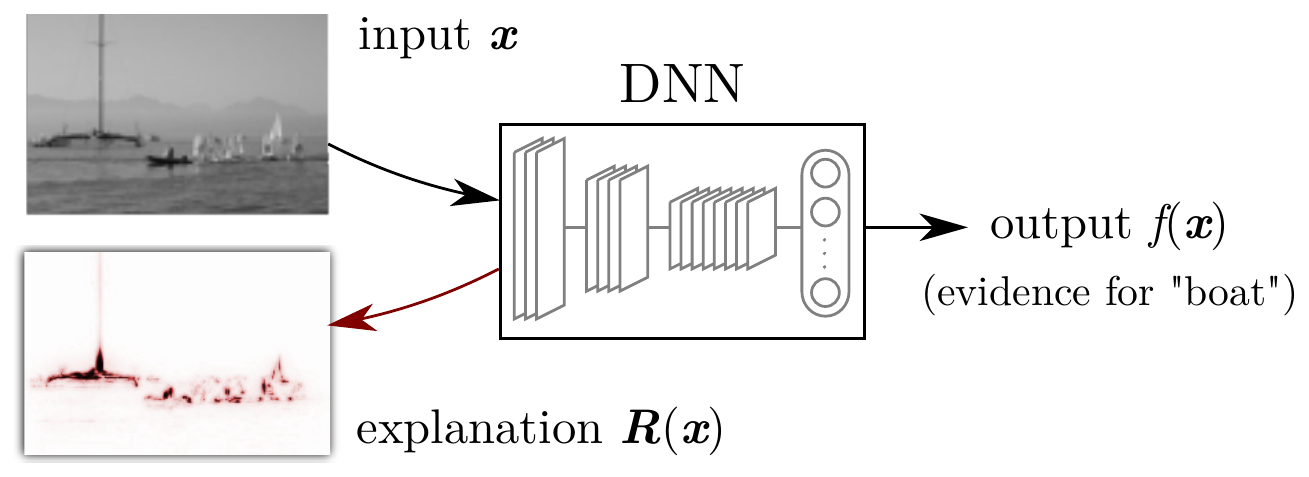}
\vskip -1mm
\caption{Explanation of the DNN prediction ``boat'' for an image $\x$ given as input.}
\label{figure:explanation}
\end{figure}

In this section, we ask for a given data point $\x$, what makes it representative of a certain concept $\omega_c$ encoded in some output neuron of the deep neural network (DNN). The output neuron can be described as a function $f(\x)$ of the input. A common approach is to view the data point $\x$ as a collection of features $(x_i)_{i=1}^d$, and to assign to each of these, a score $R_i$ determining how {\em  relevant} the feature $x_i$ is for explaining $f(\x)$. An example is given in Figure~\ref{figure:explanation}.

In this example, an image is presented to a DNN and is classified as ``boat''. The prediction (encoded in the output layer) is then mapped back to the input domain. The explanation takes the form of a heatmap, where pixels with a high associated relevance score are shown in red.

\subsection{Sensitivity Analysis}
\label{section:explanation-sensitivity}

A first approach to identify the most important input features is sensitivity analysis. It is based on the model's locally evaluated gradient or some other local measure of variation. A common formulation of sensitivity analysis defines relevance scores as
$$
R_i(\x) = \Big(\frac{\partial f}{\partial x_i}\Big)^2,
$$
where the gradient is evaluated at the data point $\x$. The most relevant input features are those to which the output is most sensitive. The technique is easy to implement for a deep neural network, since the gradient can be computed using backpropagation \citep{Bishop:1995:NNP:525960, Rumelhart1986}.

Sensitivity analysis has been regularly used in scientific applications of machine learning such as medical diagnosis \citep{Khan2001}, ecological modeling \citep{Gevrey2003}, or mutagenicity prediction \citep{DBLP:journals/jmlr/BaehrensSHKHM10}. More recently, it was also used for explaining the classification of images by deep neural networks~\citep{DBLP:journals/corr/SimonyanVZ13}.

It is important to note, however, that sensitivity analysis does not produce an explanation of the function value $f(\x)$ itself, but rather a {\em variation} of it. Sensitivity scores are indeed a decomposition of the local variation of the function as measured by the gradient square norm:
$$
\textstyle \sum_{i=1}^d R_i(\x) = \|\nabla f(\x)\|^2
$$
Intuitively, when applying sensitivity analysis e.g.\ to a neural network detecting cars in images, we answer the question ``{\em what makes this image more/less a car?}'', rather than the more basic question ``{\em what makes this image a car?}''.

\subsection{Simple Taylor Decomposition}
\label{section:explanation-taylor}

The Taylor decomposition \citep{Bazen2013,10.1371/journal.pone.0130140} is a method that explains the model's decision by decomposing the function value $f(\x)$ as a sum of relevance scores. The relevance scores are obtained by identification of the terms of a first-order Taylor expansion of the function at some root point $\widetilde \x$ for which $f(\widetilde \x) = 0$. This expansion lets us rewrite the function as:
$$
\textstyle f(\x) = \sum_{i=1}^d R_i(\x) + O(\x\x^\top)
$$
where the relevance scores
$$
R_i(\x) = \frac{\partial f}{\partial x_i}\Big|_{\x = \widetilde \x} \cdot (x_i - \widetilde x_i)
$$
are the first-order terms, and where $O(\x\x^\top)$ contains all higher-order terms. Because these higher-order terms are typically non-zero, this analysis only provides a partial explanation of $f(\x)$.

However, a special class of functions, piecewise linear and satisfying the property $f(t\,\x) = t\,f(\x)$ for $t \geq 0$, is not subject to this limitation. Examples of such functions used in machine learning are homogeneous linear models, or deep ReLU networks (without biases). For these functions, we can always find a root point $\widetilde \x = \lim_{\varepsilon \to 0} \varepsilon \cdot \x$, that incidentally lies on the same linear region as the data point $\x$, and for which the second and higher-order terms are zero. In that case, the function can be rewritten as
$$
\textstyle f(\x) = \sum_{i=1}^d R_i(\x)
$$
where the relevance scores simplify to
$$
R_i(\x) = \frac{\partial f}{\partial x_i} \cdot x_i.
$$
Relevance can here be understood as the product of sensitivity (given by the locally evaluated partial derivative) and saliency (given by the input value). That is, an input feature is relevant if it is both present in the data, and if the model reacts to it.

Later in this tutorial, we will also show how this simple technique serves as a primitive for building the more sophisticated deep Taylor decomposition~\citep{DBLP:journals/pr/MontavonLBSM17}.

\subsection{Relevance Propagation}
\label{section:explanation-propagation}

An alternative way of decomposing the prediction of a DNN is to make explicit use of its feed-forward graph structure. The algorithm starts at the output of the network, and moves in the graph in reverse direction, progressively redistributing the prediction score (or total relevance) until the input is reached. The redistribution process must furthermore satisfy a local \emph{relevance conservation} principle.

A physical analogy would be that of an electrical circuit where one injects a certain amount of current at the first endpoint, and measures the resulting current at the other endpoints. In this physical example, Kirchoff's conservation laws for current apply locally to each node of the circuit, but also ensure the conservation property at a global level.

The propagation approach was proposed by \citet{DBLP:conf/cidm/LandeckerTBMKB13} to explain the predictions of hierarchical networks, and was also introduced by \citet{10.1371/journal.pone.0130140} in the context of convolutional DNNs for explaining the predictions of these state-of-the-art models.

Let us consider a DNN where $j$ and $k$ are indices for neurons at two successive layers. Let $(R_k)_k$ be the relevance scores associated to neurons in the higher layer. We define $R_{j \leftarrow k}$ as the share of relevance that flows from neuron $k$ to neuron $j$. This share is determined based on the contribution of neuron $j$ to $R_k$, subject to the local relevance conservation constraint
$$
\textstyle \sum_j R_{j \leftarrow k} = R_k.
$$
The relevance of a neuron in the lower layer is then defined as the total relevance it receives from the higher layer:
$$
\textstyle R_j = \sum_{k} R_{j \leftarrow k}
$$
These two equations, when combined, ensure between all consecutive layers a relevance conservation property, which in turn also leads to a global conservation property from the neural network output to the input relevance scores:
$$
\textstyle \sum_{i=1}^d R_i = \dots = \sum_j R_j = \sum_k R_k = \dots = f(\x)
$$

It should be noted that there are other explanation techniques that rely on the DNN graph structure, although not producing a decomposition of $f(\x)$. Two examples are the \emph{deconvolution} by \citet{DBLP:conf/eccv/ZeilerF14}, and \emph{guided backprop} by \citet{DBLP:journals/corr/SpringenbergDBR14}. They also work by applying a backward mapping through the graph, and generate interpretable patterns in the input domain, that are associated to a certain prediction or a feature map activation.

\subsection{Practical Considerations}

Explanation techniques that derive from a decomposition principle provide several practical advantages: First, they give an implicit quantification of the share that can be imputed to individual input features. When the number of input variables is limited, the analysis can therefore be represented as a pie chart or histogram. If the number of input variables is too large, the decomposition can be coarsened by {\em pooling} relevance scores over groups of features.

For example, in RGB images, the three relevance scores of a pixel can be summed to obtain the relevance score of the whole pixel. The resulting pixel scores can be displayed as a heatmap. On an object recognition task, \citet{DBLP:conf/cvpr/LapuschkinBMMS16} further exploited this mechanism by pooling relevance over two large regions of the image: (1) the bounding box of the object to detect and (2) the rest of the image. This coarse analysis was used to quantify the reliance of the model on the object itself and on its spatial context.

In addition, when the explanation technique uses propagation in the model's graph, the quantity being propagated can be {\em filtered} to only include what flows through a certain neuron or feature map. This allows to capture individual components of an explanation, that would otherwise be entangled in the heatmap.

The pooling and filtering capabilities of each explanation technique are shown systematically in Table~\ref{table:comparison}.
\begin{table}[h!]
\centering { \small
\begin{tabular}{|l|cc|}\hline
 & pooling & filtering\\\hline
sensitivity analysis & $\checkmark$  & \\\hline
simple Taylor & $\checkmark$  & \\\hline
relevance propagation & $\checkmark$ & $\checkmark$\\\hline
{\em deconvolution} \citep{DBLP:conf/eccv/ZeilerF14} & & $\checkmark$ \\\hline
{\em guided backprop} \citep{DBLP:journals/corr/SpringenbergDBR14} & & $\checkmark$ \\\hline
\end{tabular}
}
\caption{Properties of various techniques for explaining DNN decisions. The first three entries correspond to the methods introduced in Sections \ref{section:explanation-sensitivity}--\ref{section:explanation-propagation}.}
\label{table:comparison}
\end{table}

\section{The LRP Explanation Framework}
\label{section:lrp}
In this section, we focus on the layer-wise relevance propagation (LRP) technique introduced by \citet{10.1371/journal.pone.0130140} for explaining deep neural network predictions. LRP is based on the propagation approach described in Section \ref{section:explanation-propagation}, and has been used in a number of practical applications, in particular, for model validation and analysis of scientific data. Some of these applications are discussed in Sections \ref{section:validation} and~\ref{section:scientific}.

LRP is first described algorithmically in Section \ref{section:lrp-rules}, and then shown in Section \ref{section:lrp-deeptaylor} to correspond in some cases to a deep Taylor decomposition of the model's decision \citep{DBLP:journals/pr/MontavonLBSM17}. Practical recommendations and tricks to make efficient use of LRP are then given in Section~\ref{section:tricks}.

\subsection{Propagation Rules for DNNs}
\label{section:lrp-rules}

In the original paper \citep{10.1371/journal.pone.0130140}, LRP was applied to bag-of-words and deep neural network models. In this tutorial, we focus on the second type of models. Let the neurons of the DNN be described by the equation
$$
\textstyle a_k = \sigma\big(\sum_j a_j w_{jk} + b_k\big),
$$
with $a_k$ the neuron activation, $(a_j)_j$ the activations from the previous layer, and $w_{jk},b_k$ the weight and bias parameters of the neuron. The function $\sigma$ is a positive and monotonically increasing activation function.

One propagation rule that fulfills local conservation properties, and that was shown to work well in practice is the $\alpha\beta$-rule given by:
\begin{align}
R_j = \sum_{k} \Big(\alpha \frac{a_j w_{jk}^+}{\sum_j a_j w_{jk}^+} - \beta \frac{a_j w_{jk}^-}{\sum_j a_j w_{jk}^-}\Big)  R_k,
\label{eq:lrp-alphabeta}
\end{align}
where $()^+$ and $()^-$ denote the positive and negative parts respectively, and where the parameters $\alpha$ and $\beta$ are chosen subject to the constraints $\alpha-\beta=1$ and $\beta \geq 0$. To avoid divisions by zero, small stabilizing terms can be introduced when necessary. The rule can be rewritten as
$$
R_j = \sum_{k} \frac{a_j w_{jk}^+}{\sum_j a_j w_{jk}^+} R_k^{\wedge} + \sum_{k} \frac{a_j w_{jk}^-}{\sum_j a_j w_{jk}^-}  R_k^{\vee},
$$
where $R_k^{\wedge} = \alpha R_k$ and $R_k^{\vee} = -\beta R_k$. It can now be interpreted as follows:
\begin{quote}
\em Relevance $R_k^{\wedge}$ should be redistributed to the lower-layer neurons $(a_j)_j$ in proportion to their excitatory effect on $a_k$. ``Counter-relevance'' $R_k^{\vee}$ should be redistributed to the lower-layer neurons $(a_j)_j$ in proportion to their inhibitory effect on $a_k$. 
\end{quote}
Different combinations of parameters $\alpha,\beta$ were shown to modulate the qualitative behavior of the resulting explanation. As a naming convention, we denote, for example, by LRP-$\alpha_2\beta_1$, the fact of having chosen the parameters $\alpha=2$ and $\beta=1$ for this rule. In the context of image classification, a non-zero value for $\beta$ was shown empirically to have a sparsifying effect on the explanation \citep{10.1371/journal.pone.0130140,DBLP:journals/pr/MontavonLBSM17}. On the BVLC CaffeNet \citep{DBLP:conf/mm/JiaSDKLGGD14}, LRP-$\alpha_2\beta_1$ was shown to work well, while for the deeper GoogleNet \citep{DBLP:conf/cvpr/SzegedyLJSRAEVR15}, LRP-$\alpha_1\beta_0$ was found to be more stable.

When choosing LRP-$\alpha_1\beta_0$, the propagation rule reduces to the simpler rule:
\begin{align}
R_j = \sum_{k} \frac{a_j w_{jk}^+}{\sum_j a_j w_{jk}^+} R_k.
\label{eq:alpha}
\end{align}
The latter rule was also used by \citet{DBLP:conf/eccv/ZhangLBSS16} as part of an explanation method called excitation backprop.

\subsection{LRP and Deep Taylor Decomposition}
\label{section:lrp-deeptaylor}

In this section, we show for deep ReLU networks a connection between LRP-$\alpha_1\beta_0$ and Taylor decomposition. We show in particular that when neurons are defined as
$$
\textstyle a_k = \max\big(0,\sum_j a_j w_{jk} + b_k\big) \quad \text{with} \quad b_k \leq 0,
$$
the application of LRP-$\alpha_1\beta_0$ at a given layer can be seen as computing a Taylor decomposition of the relevance at that layer onto the lower layer. The name ``deep Taylor decomposition'' then arises from the iterative application of Taylor decomposition from the top layer down to the input layer.

\begin{figure}[h!]
\centering
\includegraphics[width=1.0\linewidth]{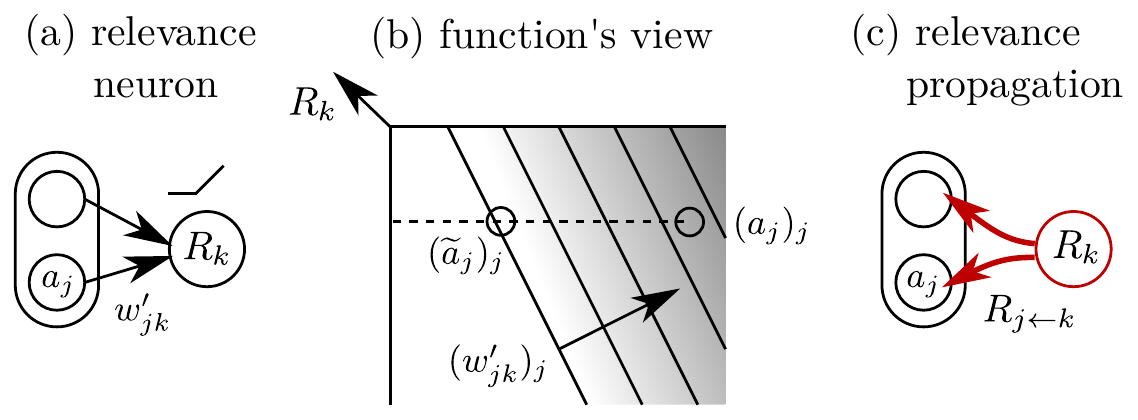}
\caption{Diagram of the relevance neuron and its analysis. The root search domain is shown with a dashed line, and the relevance propagation resulting from decomposing $R_k$ is shown in red.}
\label{figure:relu}
\end{figure}

Let us assume that the relevance for the neuron $k$ can be written as $R_k = a_k c_k$, a product of the neuron activation $a_k$ and a term $c_k$ that is {\em constant} and {\em positive}. These two properties allow us to construct a ``relevance neuron''
\begin{align}
\textstyle R_k = \max\big(0,\sum_j a_j w_{jk}^\prime + b_k^\prime \big),
\label{eq:relneuron}
\end{align}
with parameters $w_{jk}^\prime = w_{jk} c_{k}$ and $b_k^\prime = b_k c_{k}$. The relevance neuron is shown in Figure~\ref{figure:relu}(a).

We now would like to propagate the relevance to the lower layer. For this, we perform a Taylor decomposition of $R_k$ on the lower-layer neurons. We search for the nearest root point $(\widetilde a_j)_j$ of $R_k$ on the segment $[(a_j 1_{w_{jk}^\prime \leq 0})_j ,(a_j)_j]$. The search strategy is visualized in Figure~\ref{figure:relu}(b). Because the relevance neuron is piecewise linear, the Taylor expansion at the root point contains only first-order terms:
$$
R_k = \sum_j \underbrace{\frac{\partial R_k}{\partial a_j} \Big|_{(\widetilde a_j)_j}\!\! \cdot (a_j - \widetilde a_j)}_{R_{j \leftarrow k}}
$$
The first-order terms correspond to the decomposition of $R_k$ on the lower-layer neurons and have the closed-form expression
$$
R_{j \leftarrow k} = \frac{a_j w_{jk}^+ }{ \sum_j a_j w_{jk}^+} R_k.
$$
The resulting propagation of $R_k$ is shown in Figure~\ref{figure:relu}(c). Summing $R_{j \leftarrow k}$ over all neurons $k$ to which neuron $j$ contributes yields exactly the LRP-$\alpha_1\beta_0$ propagation rule of Equation~\eqref{eq:alpha}.

We now would like to verify that the procedure can be repeated one layer below. For this, we inspect the structure of $R_j$ and observe that it can be written as a product $R_j = a_j c_j$, where $a_j$ is the neuron activation and
\begin{align*}
c_j &= \sum_{k}  \frac{w_{jk}^+}{\sum_j a_j w_{jk}^+} R_k\\
&= \sum_{k} w_{jk}^+ \frac{\max\big(0,\sum_j a_j w_{jk}+b_k\big)}{\sum_j a_j w_{jk}^+} c_k
\end{align*}
is {\em positive} and also approximately {\em constant}. The latter property arises from the observation that the dependence of $c_j$ on the activation $a_j$ is only very indirect (diluted by two nested sums), and that the other terms $w_{jk},w_{jk}^+,b_k,c_k$ are constant or approximately constant.

The positivity and near-constancy of $c_j$ implies that similar relevance neuron to the one of Equation~\eqref{eq:relneuron} can be built for neuron $j$, for the purpose of redistributing relevance on the layer before. The decomposition process can therefore be repeated in the lower layers, until the first layer of the neural network is reached, thus, performing a deep Taylor decomposition~\citep{DBLP:journals/pr/MontavonLBSM17}.

In the derivation above, the segment on which we search for a root point incidentally guarantees (1) membership of the root point to the domain of ReLU activations and (2) positivity of relevance scores. These guarantees can also be brought to other types of layers (e.g.\ input layers receiving real values or pixels intensities), by searching for a root point $(\widetilde a_j)_j$ on a different segment. This leads to different propagation rules, some of which are listed in Table \ref{table:deeptaylor}. Details on how to derive these rules are given in the original paper \citep{DBLP:journals/pr/MontavonLBSM17}. We refer to these rules as ``deep Taylor LRP'' rules.

\begin{table}[h!]
\centering \small
\def\arraystretch{1.3}
\begin{tabular}{|l|l|}\hline
\parbox{2.5cm}{Input domain} & Rule \\\hline
\parbox{2.5cm}{\vskip 1mm ReLU activations\\($a_j \geq 0$)\vskip 1mm } &
$\displaystyle R_j = \sum_{k} \frac{a_j w_{jk}^+}{\sum_j a_j w_{jk}^+} R_k$
\\\hline
\parbox{2.5cm}{\vskip 1mm Pixel intensities\\($x_i \in [l_i,h_i]$,\\$l_i \leq 0 \leq h_i$)\vskip 1mm} &
$\displaystyle R_i = \sum_{j} \frac{x_i w_{ij} - l_i w_{ij}^+ - h_i w_{ij}^-}{\sum_i x_i w_{ij} - l_i w_{ij}^+ - h_i w_{ij}^-} R_j$
\\\hline
\parbox{2.5cm}{\vskip 1mm Real values\\($x_i \in \mathbb{R}$)\vskip 1mm } &
$\displaystyle R_i = \sum_{j} \frac{w_{ij}^2}{\sum_i w_{ij}^2} R_j$
\\\hline
\end{tabular}
\caption{Deep Taylor LRP rules derived for various layer types. The first rule applies to the hidden layers, and the next two rules apply to the first layer.}
\label{table:deeptaylor}
\end{table}

\subsection{Handling Special Layers}

Practical neural networks are often composed of special layers, for example, $\ell_p$-pooling layers (including sum-pooling and max-pooling as the two extreme cases), and normalization layers. The original paper by \citet{10.1371/journal.pone.0130140} uses a winner-take-all redistribution policy for max-pooling layers, where all relevance goes to the most activated neuron in the pool. Instead, \citet{DBLP:journals/pr/MontavonLBSM17} recommend to apply for $\ell_p$-pooling layers the following propagation rule:
$$
R_j = \frac{x_j}{\sum_j x_j} R_k,
$$
i.e.\ redistribution is proportional to neuron activations in the pool. This redistribution rule ensures explanation continuity (see Section \ref{section:quality-continuity} for an introduction to this concept). 

With respect to normalization layers, \citet{10.1371/journal.pone.0130140} proposed to ignore them in the relevance propagation pass. Alternately, \citet{DBLP:conf/icann/BinderMLMS16} proposed for these layers a more sophisticated rule based on a local Taylor expansion of the normalization function, with some benefits in terms of explanation selectivity. 

\section{Recommendations and Tricks for LRP}
\label{section:tricks}

Machine learning methods are often described in papers at an abstract level, for maximum generality. However, a good choice of hyperparameters is usually necessary to make them work well on real-world problems, and tricks are often used to make most efficient use of these methods and extend their capabilities \citep{DBLP:series/lncs/Bengio12,DBLP:series/lncs/Hinton12,Montavon:2012:NNT:2480981}. Likewise, the LRP framework introduced in Section \ref{section:lrp}, also comes with a list of recommendations and tricks, some of which are given below.

\subsection{How to Choose the Model to Explain}

The LRP approach is aimed at general feedforward computational graphs. However, it was most thoroughly studied, both theoretically \citep{DBLP:journals/pr/MontavonLBSM17} and empirically \citep{Samek2016}, on specific types of models such as convolutional neural networks with ReLU nonlinearities. This leads to our first recommendation:

\begin{quote} \em Apply LRP to classes of models where it was successfully applied in the past. In absence of trained model of such class, consider training your own.
\end{quote}

We have also observed empirically that in order for LRP to produce good explanations, the number of fully connected layers should be kept low, as LRP tends for these layers to redistribute relevance to too many lower-layer neurons, and thus, loose selectivity.

\begin{quote} \em As a first try, consider a convolutional ReLU network, as deep as needed, but with not too many fully connected layers. Use dropout {\rm \citep{DBLP:journals/jmlr/SrivastavaHKSS14}} in these layers.\end{quote}

For the LRP procedure to best match the deep Taylor decomposition framework outlined in Section \ref{section:lrp-deeptaylor}, sum-pooling or average-pooling layers should be preferred to max-pooling layers, and bias parameters of the network should either be zero or negative.

\begin{quote} \em Prefer sum-pooling to max-pooling, and force biases to be zero or negative at training time.
\end{quote}

Negative biases will contribute to further sparsify the network activations, and therefore, also to better disentangle the relevance at each layer.

\subsection{How to Choose the LRP Rules for Explanation}

In presence of a deep neural network that follows the recommendations above, a first set of propagation rules to be tried are the deep Taylor LRP rules of Table \ref{table:deeptaylor}, which exhibit a stable behavior, and that are also well understood theoretically. These rules produce for positive predictions a positive heatmap, where input variables are deemed relevant if $R_i > 0$ or irrelevant if $R_i = 0$.

\begin{quote} \em As a default choice for relevance propagation, use the deep Taylor LRP rules given in Table~\ref{table:deeptaylor}.
\end{quote}

In presence of predictive uncertainty, a certain number of input variables might be in contradiction with the prediction, and the concept of ``negative relevance'' must therefore be introduced. Negative relevance can be injected into the explanation in a controlled manner by setting the coefficients of the $\alpha\beta$-rule of Equation~\eqref{eq:lrp-alphabeta} to an appropriate value.

\begin{quote} \em 
If negative relevance is needed, or the heatmaps are too diffuse, replace the rule LRP-$\alpha_1\beta_0$ by LRP-$\alpha_2\beta_1$ in the hidden layers.
\end{quote}

The LRP-$\alpha_1\beta_0$ and LRP-$\alpha_2\beta_1$ rules were shown to work well on image classification \citep{DBLP:journals/pr/MontavonLBSM17}, but there is a potentially much larger set of rules that we can choose from. For example, the ``$\epsilon$-rule'' \citep{10.1371/journal.pone.0130140} was applied successfully to text categorization \citep{DBLP:journals/corr/ArrasHMMS16a,DBLP:journals/corr/ArrasMMS17a}. To choose the most appropriate rule among the set of possible ones, a good approach is to define a heatmap quality criterion, and select the rule at each layer accordingly. One such quality criterion called ``pixel-flipping'' measures heatmap selectivity and is later introduced in Section~\ref{section:quality-selectivity}.

\begin{quote} \em 
If the heatmaps obtained with LRP-$\alpha_1\beta_0$ and LRP-$\alpha_2\beta_1$ are unsatisfactory, consider a larger set of propagation rules, and use pixel-flipping to select the best one.
\end{quote}

\subsection{Tricks for Implementing LRP}

Let us consider the LRP-$\alpha_1\beta_0$ propagation rule of Equation~\eqref{eq:alpha}:
$$
R_j = a_j \sum_{k} \frac{ w_{jk}^+}{\sum_j a_j w_{jk}^+} R_k,
$$
where we have for convenience moved the neuron activation $a_j$ outside the sum. This rule can be written as four elementary computations, all of which can also expressed in vector form:
\begin{align}
&\text{element-wise} & &\text{vector form} \nonumber\\\hline \nonumber\\[-5mm]
z_k            &\leftarrow \textstyle \sum_j a_j w_{jk}^+ &
\z             &\leftarrow W_+^\top \cdot \boldsymbol{a} \label{eq:lrpvec-a}\\
s_k            &\leftarrow \textstyle R_k / z_k &
\boldsymbol{s} &\leftarrow \boldsymbol{R} \oslash \z \label{eq:lrpvec-b}\\
c_j &\leftarrow \textstyle \sum_k w_{jk}^+ s_k &
\boldsymbol{c} &\leftarrow W_+ \cdot \boldsymbol{s} \label{eq:lrpvec-c}\\
R_j &\leftarrow \textstyle a_j c_j &
\boldsymbol{R} &\leftarrow \boldsymbol{a} \odot \boldsymbol{c} \label{eq:lrpvec-d}\\\hline\nonumber
\end{align}
\vskip -5mm
In the vector form computations, $\oslash$ and $\odot$ denote the element-wise division and multiplication. The variable $W$ denotes the weight matrix connecting the neurons of the two consecutive layers, and $W_+$ is the matrix retaining only the positive weights of $W$ and setting remaining weights to zero. This vector form is useful to implement LRP for fully connected layers.

In convolution layers, the matrix-vector multiplications of Equations~\eqref{eq:lrpvec-a} and~\eqref{eq:lrpvec-c} can be more efficiently implemented by borrowing the \texttt{forward} and \texttt{backward} methods used for forward activation and gradient propagation. These methods are readily available in many neural network libraries and are typically highly optimized. Based on these high-level primitives, LRP can implemented by the following sequence of operations:

\vbox{
\vskip 2mm \hrule \vskip 2mm
{\ttfamily
\noindent \phantom{--}{\fontseries{b}\selectfont def} lrp(layer,a,R):\\[2mm]
\phantom{------}clone = layer.clone()\\
\phantom{------}clone.W = maximum(0,layer.W)\\
\phantom{------}clone.B = 0\\[2mm]
\phantom{------}z = clone.forward(a)\\
\phantom{------}s = R / z\\
\phantom{------}c = clone.backward(s)\\[2mm]
\phantom{------}{\fontseries{b}\selectfont return} a * c
}
\vskip 2mm \hrule \vskip 2mm
}

The function \texttt{lrp} receives as arguments the \texttt{layer} through which the relevance should be propagated, the activations ``\texttt{a}'' at the layer input, and the relevance scores ``\texttt{R}'' at the layer output. The function returns the redistributed relevance at the layer input. Sample code is provided at \texttt{http://heatmapping.org/tutorial}. This modular approach was also used by \citet{DBLP:conf/eccv/ZhangLBSS16} to implement the excitation backprop method.

\subsection{Translation Trick for Denoising Heatmaps}

It is sometimes observed that, for classifiers that are not optimally trained or structured, LRP heatmaps have unaesthetic features. This can be caused, for example, by the presence of noisy first-layer filters, or a large stride parameter in the first convolution layer. These effects can be mitigated by considering the explanation not of a single input image but the explanations of multiple slightly translated versions of the image. The heatmaps for these translated versions are then recombined by applying to them the inverse translation operation and averaging them up. In mathematical terms, the improved heatmap is given by:
$$
\boldsymbol{R}^\star(\x) = \frac{1}{|\mathcal{T}|}\sum_{\tau \in \mathcal{T}} \tau^{-1}(\boldsymbol{R}(\tau(\x)))
$$
where $\tau,\tau^{-1}$ denote the translation and its inverse, and $\mathcal{T}$ is the set of all translations of a few pixels.

\subsection{Sliding Window Explanations for Large Images}

\begin{figure}[h!]
\centering
\includegraphics[width=0.85\linewidth]{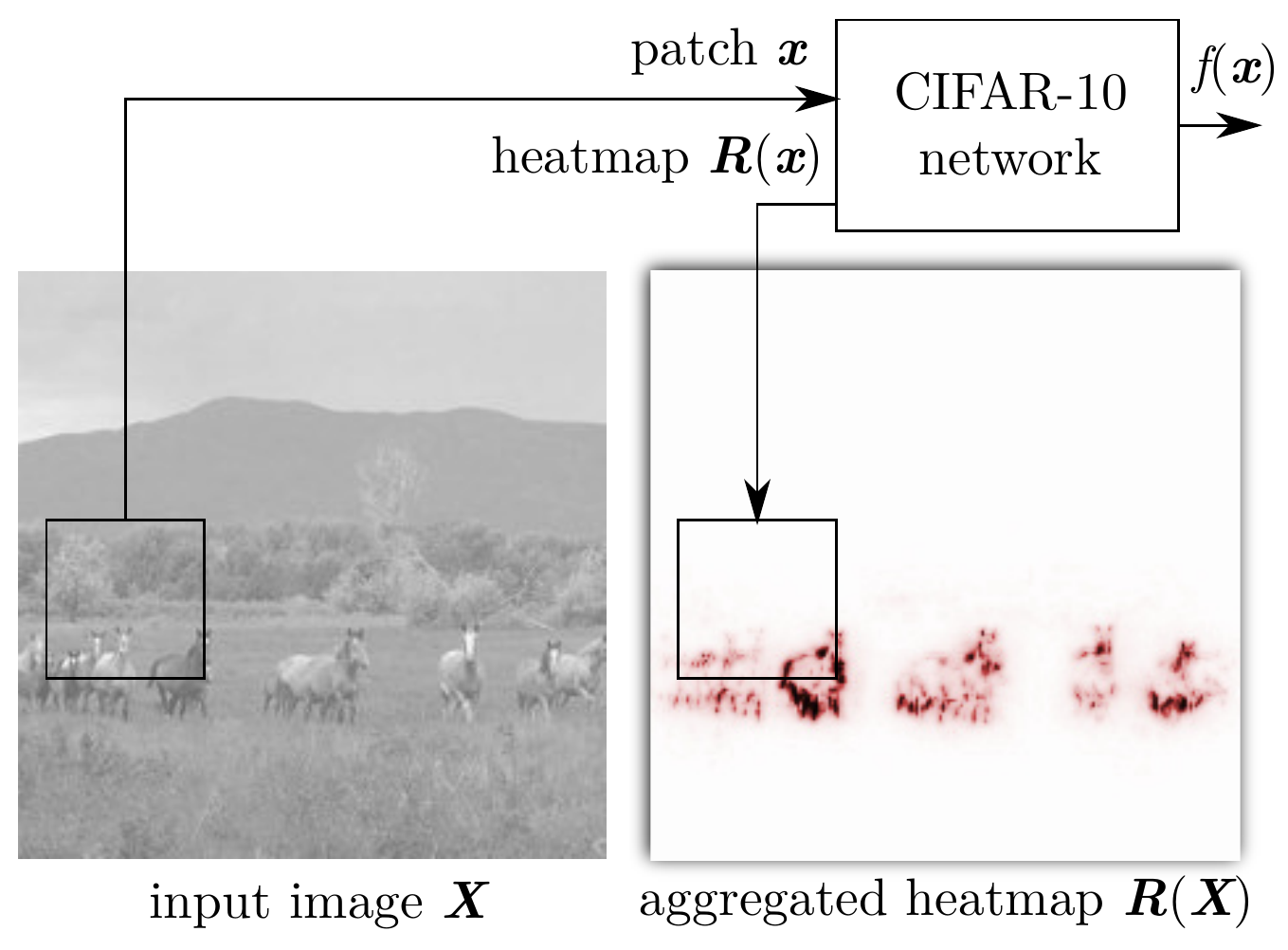}
\vskip -1mm
\caption{Highlighting in a large image pixels that are relevant for the CIFAR-10 class ``horse'', using the sliding window technique.}
\label{figure:convolved}
\end{figure}

In applications such as medical imaging or scene parsing, the images to be processed are typically larger than the what the neural network receives as input. Let $\boldsymbol{X}$ be this large image. The LRP procedure can be extended for this scenario by applying a sliding window strategy, where the neural network is moved through the whole image, and where heatmaps produced at various locations must then be combined into a single large heatmap. Technically, we define the quantity to explain as:
$$
g(\boldsymbol{X}) = \sum_{s \in \mathcal{S}} f(\underbrace{\boldsymbol{X}[s]}_{\x})
$$
where $\boldsymbol{X}[s]$ extracts a patch from the image $\boldsymbol{X}$ at location $s$, and $\mathcal{S}$ is the set of all locations in that image. Pixels then receive relevance from all patches to which they belong and in which they contribute to the function value $f(\x)$. This technique is illustrated in Figure~\ref{figure:convolved}.

The convolutional neural network is a special case that can technically receive an input of any size. A heatmap can be obtained directly from it by redistributing the top-layer activations using LRP. This direct approach can provide a computational gain compared to the sliding window approach. However, it is not strictly equivalent and can produce unreliable heatmaps, e.g. when the network uses border-padded convolutions. If in doubt, it is preferable to use the sliding window formulation.

\subsection{Visualize Relevant Pattern}

Due to their characteristic spatial structure, LRP heatmaps readily provide intuition on which input pattern the model has used to make its prediction. However, in presence of cluttered scenes, a better visualization can be obtained by using the heatmap as a mask to extract relevant pixels (and colors) from the image. We call the result of the masking operation the \emph{pattern} $\boldsymbol{P}(\x)$ that we compute as:
$$
\boldsymbol{P}(\x) = \x \odot \boldsymbol{R}(\x).
$$
Here, we assume that the heatmap scores have been preliminarily normalized between $0$ and $1$ through rescaling and/or clipping so that the masked image remains in the original color space. This visualization of LRP heatmaps makes it also more directly comparable to the visualizations techniques proposed in \citep{DBLP:conf/eccv/ZeilerF14,DBLP:journals/corr/SpringenbergDBR14}.

\section{Quantifying Explanation Quality}

In Sections \ref{section:explanation} and \ref{section:lrp}, we have introduced a number of explanation techniques. While each technique is based on its own intuition or mathematical principle, it is also important to define at a more abstract level what are the characteristics of a good explanation, and to be able to test for these characteristics quantitatively. A quantitative framework allows to compare explanation techniques specifically for a target problem, e.g.\ ILSVRC or MIT Places \citep{Samek2016}. We present in Sections \ref{section:quality-continuity} and \ref{section:quality-selectivity} two important properties of an explanation, along with possible evaluation metrics.

\subsection{Explanation Continuity}
\label{section:quality-continuity}

\begin{figure}[h!]
\centering 
{ \small
\includegraphics[width=1.0\linewidth]{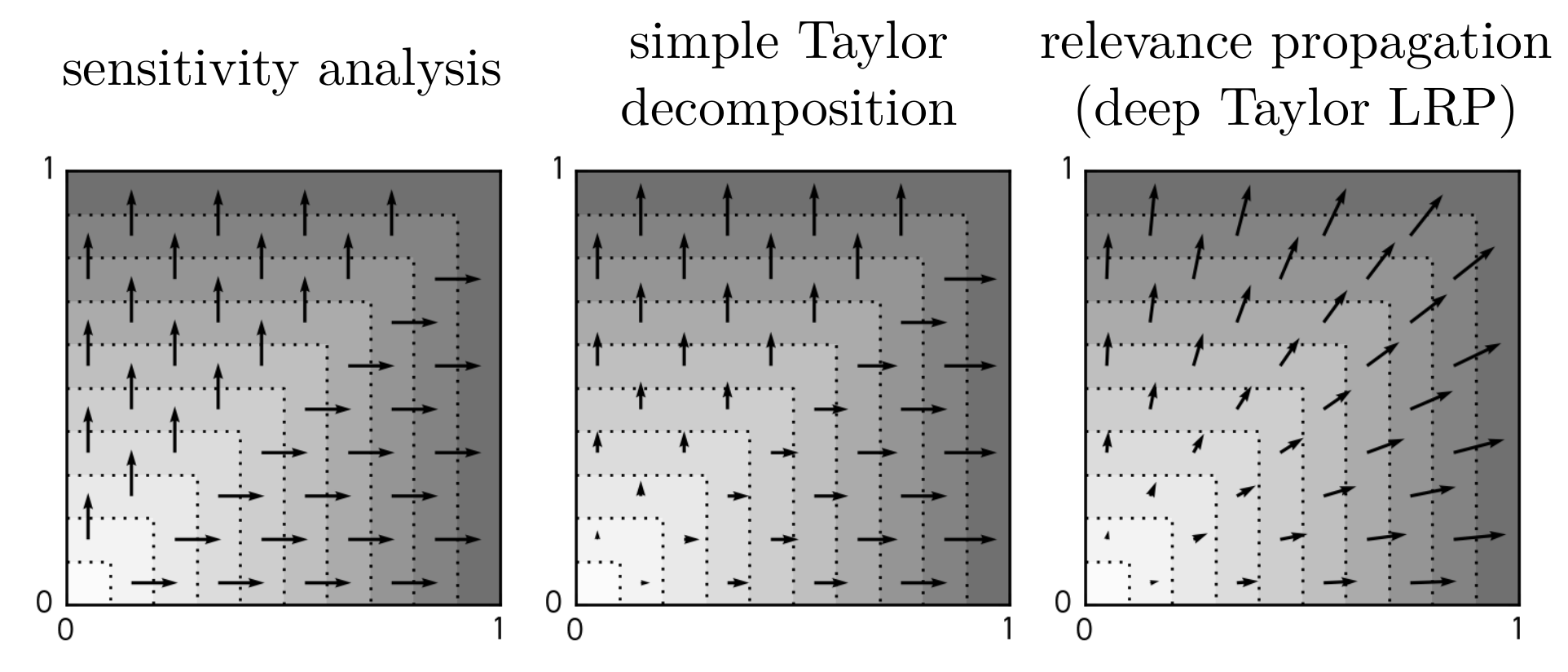}
}\vskip -1mm
\caption{Explaining $\max(x_1,x_2)$. Function values are represented as a contour plot, with dark regions corresponding to high values. Relevance scores are represented as a vector field, where horizontal and vertical components are the relevance of respective input variables.}
\label{figure:quiver}
\end{figure}

A first desirable property of an explanation technique is that it produces a continuous explanation function. Here, we implicitly assume that the prediction function $f(\x)$ is also continuous. We would like to ensure in particular the following behavior:
\begin{quote}
\em If two data points are nearly equivalent, then the explanations of their predictions should also be nearly equivalent.
\end{quote}
 Explanation continuity (or lack of it) can be quantified by looking for the strongest variation of the explanation $\boldsymbol{R}(\x)$ in the input domain:
$$
\max_{\x \neq \x'}~\frac{\|\boldsymbol{R}(\x)-\boldsymbol{R}(\x')\|_1}{\|\x-\x'\|_2}.
$$
When $f(\x)$ is a deep ReLU network, both sensitivity analysis and simple Taylor decomposition have sharp discontinuities in their explanation function. On the other hand, deep Taylor LRP produces continuous explanations. This is illustrated in Figure~\ref{figure:quiver}
for the simple function $f(\x) = \max(x_1,x_2)$ in $\mathbb{R}_+^2$, here implemented by the two-layer ReLU network
\begin{align*}
f(\x) = \max\big(0 \,,\, & 0.5 \max(0,x_1-x_2)\\
                     +\, & 0.5 \max(0,x_2-x_1)\\
                     +\, & 0.5 \max(0,x_1+x_2)\big).
\end{align*}

It can be observed that despite the continuity of the prediction function, the explanations offered by sensitivity analysis and simple Taylor decomposition are discontinuous on the line $x_1 = x_2$. Here, only deep Taylor LRP produces a smooth transition.

More generally, techniques that rely on the function's gradient, such as sensitivity analysis or simple Taylor decomposition, are more exposed to the derivative noise \citep{Snyder2012} that characterizes complex machine learning models. Consequently, these techniques are also unlikely to score well in terms of explanation continuity.

\begin{figure}[b!]
\centering \includegraphics[width=0.98\linewidth]{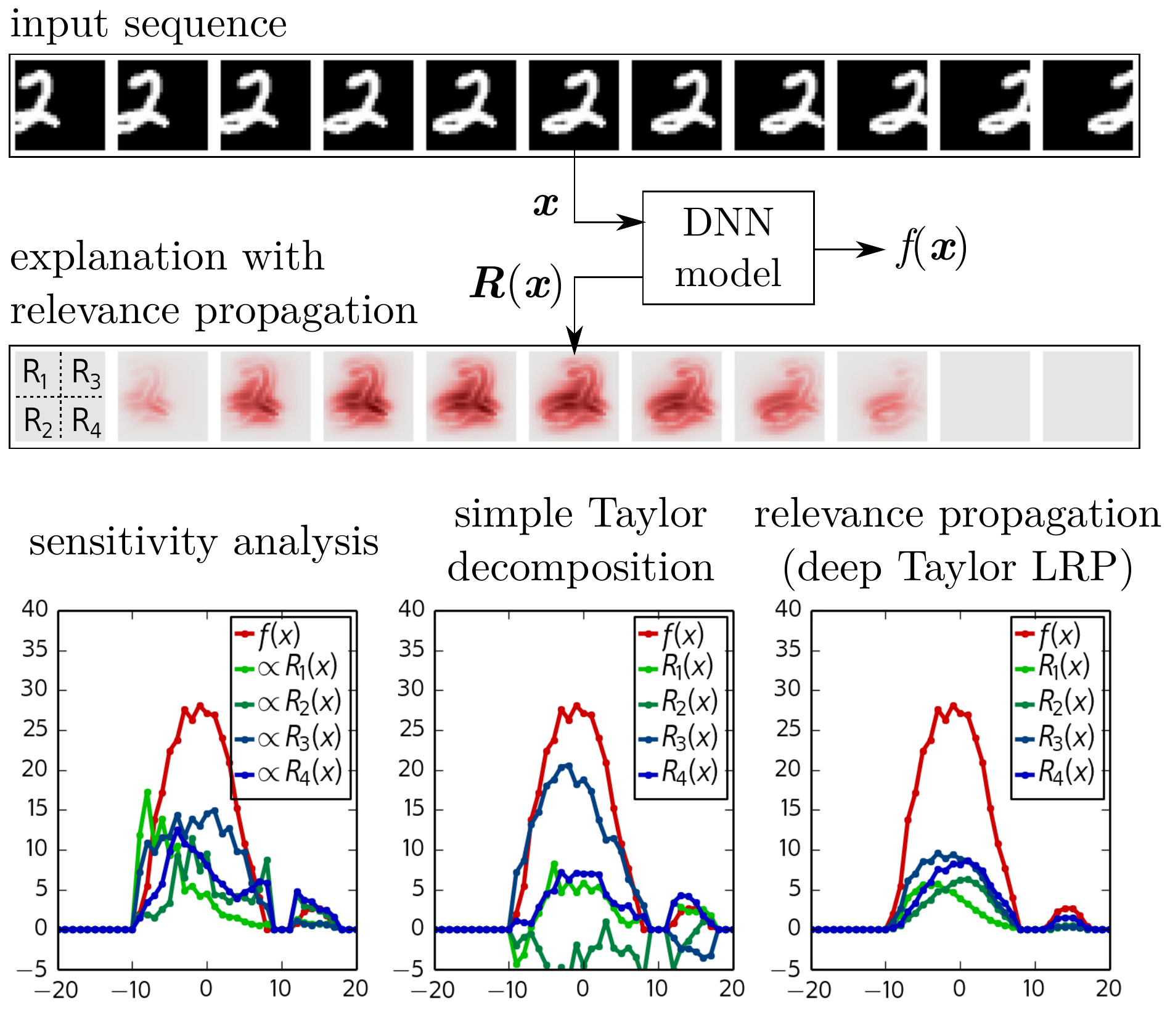}
\vskip -1mm
\caption{Classification ``2'' by a DNN, explained by different methods, as we move a handwritten digit from left to right in its receptive field. Relevance scores are pooled into four quadrants, and are tracked as we apply the translation operation.}
\label{figure:translation}
\end{figure}

Figure~\ref{figure:translation}
shows the function value and the relevance scores for each technique, when applying them to a convolutional DNN trained on MNIST. Although the function itself is relatively low-varying, strong variations occur in the explanations. Here again, only deep Taylor LRP produces reasonably continuous explanations.

\subsection{Explanation Selectivity}
\label{section:quality-selectivity}

Another desirable property of an explanation is that it redistributes relevance to variables that have the strongest impact on the function $f(\x)$. \citet{10.1371/journal.pone.0130140} and \citet{Samek2016} proposed to quantify selectivity by measuring how fast $f(\x)$ goes down when removing features with highest relevance scores.

The method was introduced for image data under the name ``pixel-flipping'' \citep{10.1371/journal.pone.0130140,Samek2016}, and was also adapted to text data, where words selected for removal have their word embeddings set to zero~\citep{DBLP:journals/corr/ArrasHMMS16a}. The method works as follows:
\begin{center}
\fbox{\parbox{0.9\linewidth}{
\textbf{repeat} until all features have been removed:
\begin{itemize}
\setlength\itemsep{0em}
\item record the current function value $f(\x)$
\item find feature $i$ with highest relevance $R_i(\x)$
\item remove that feature ($\x \leftarrow \x  - \{x_i\}$)
\end{itemize}
make a plot with all recorded function values, and return the area under the curve (AUC) for that plot.
}}
\end{center}
A sharp drop of function's value, characterized by a low AUC score indicates that the correct features have been identified as relevant. AUC results can be averaged over a large number of examples in the dataset.

\begin{figure}[b!]
\centering
\includegraphics[width=1.0\linewidth]{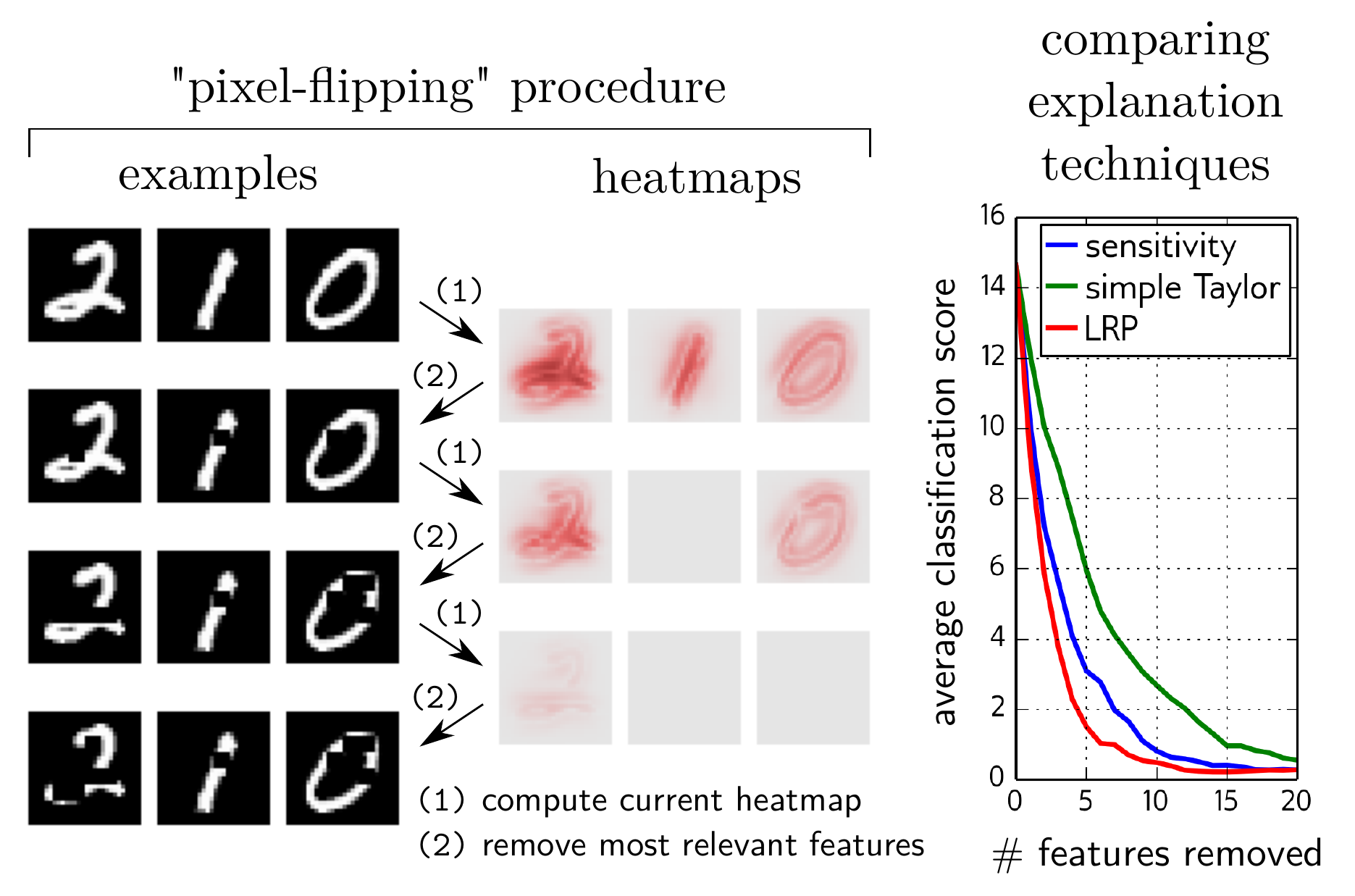}
\caption{Illustration of the ``pixel-flipping'' procedure. At each step, the heatmap is used to determine which region to remove (by setting it to black), and the classification score is recorded.}
\label{figure:pixelflip}
\end{figure}

Figure~\ref{figure:pixelflip}
illustrates the procedure on the same DNN as in Figure~\ref{figure:translation}. At each iteration, a patch of size $4 \times 4$ corresponding to the region with highest relevance is set to black. The plot on the right keeps track of the function score as the features are being progressively removed. In this particular case, the plot indicates that deep Taylor LRP is more selective than sensitivity analysis and simple Taylor decomposition.

It is important to note however, that the result of the analysis depends to some extent on the feature removal process. Various feature removal strategies can be used, but a general rule is that it should keep as much as possible the image being modified on the data manifold. Indeed, this guarantees that the DNN continues to work reliably through the whole feature removal procedure. This in turn makes the analysis less subject to uncontrolled factors of variation.

\section{Applications}

Potential applications of explanation techniques are vast and include as diverse domains as extraction of domain knowledge, computer-assisted decisions, data filtering, or compliance. We focus in this section on two types of applications: validation of a trained model, and analysis of scientific data.

\subsection{Model Validation}
\label{section:validation}

Model validation is usually achieved by measuring the error on some validation set disjoint from the training data. While providing a simple way to compare different machine learning models in practice, it should be reminded that the validation error is only a proxy for the true error and that the data distribution and labeling process might differ. A human inspection of the model rendered interpretable can be a good complement to the validation procedure. We present two recent examples showing how explainability allows to better validate a machine learning model by pointing out at some unsuspected qualitative properties of it.

\citet{DBLP:journals/corr/ArrasHMMS16a}
\begin{figure}[!b]
\centering
\includegraphics[width=1.0\linewidth]{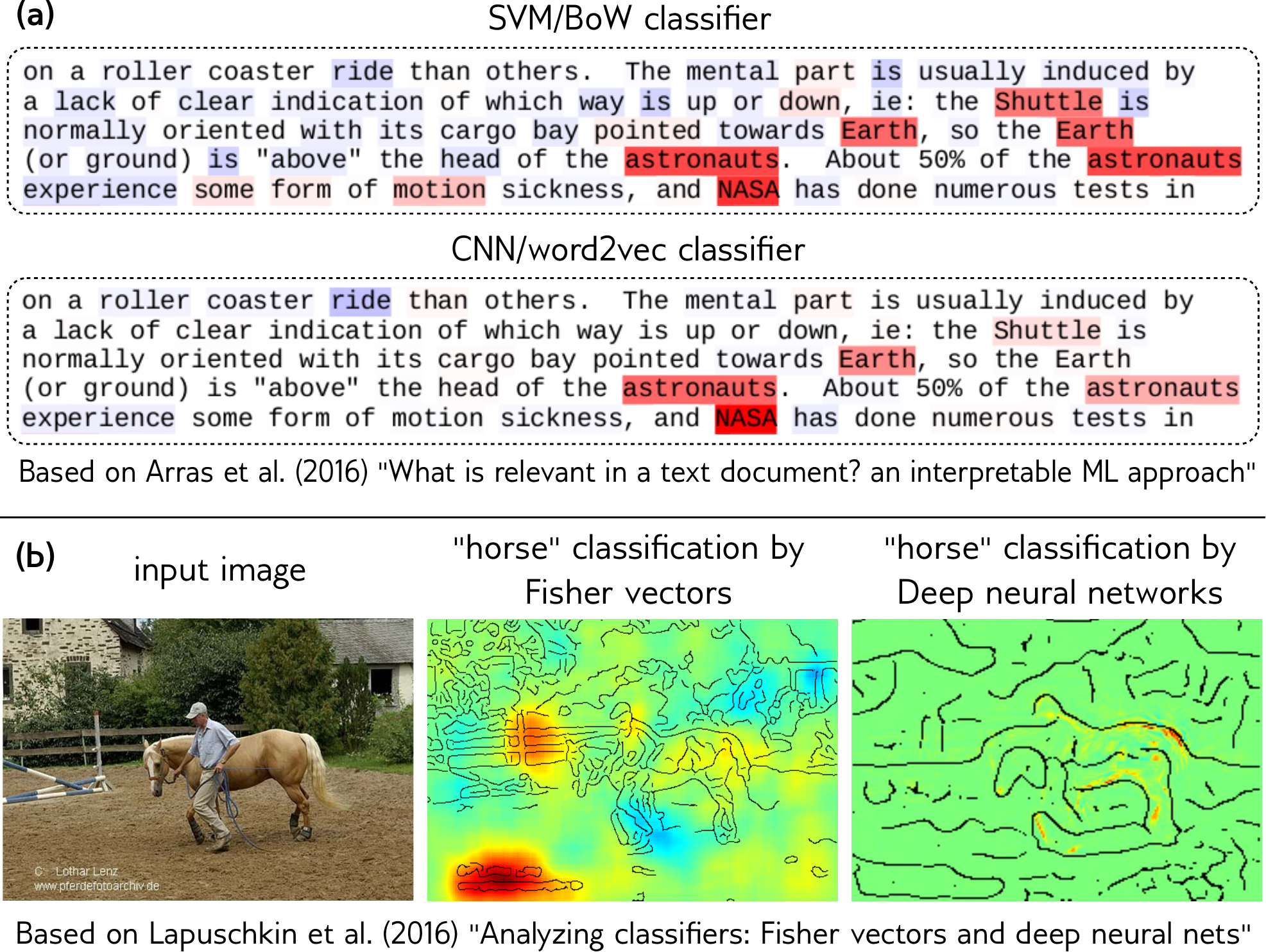}
\caption{
Examples taken from the literature of model validation via explanation. (a) Explanation of the concept ``\texttt{sci.space}'' by two text classifiers. (b) Unexpected use of copyright tags by the Fisher vector model for predicting the class ``horse''.}
\label{figure:validation}
\end{figure}
considered a document classification task on the 20-Newsgroup dataset, and compared the explanations of a convolutional neural network (CNN) trained on word2vec inputs to the explanations of a support vector machine (SVM) trained on bag-of-words (BoW) document representations. They observed that, although both models produce a similar test error, the CNN model assigns most relevance to a small number of keywords, whereas the SVM classifier relies on word count regularities. Figure~\ref{figure:validation}(a) displays explanations for an example of the target class \texttt{sci.space}.

\citet{DBLP:conf/cvpr/LapuschkinBMMS16} compared the decisions taken by convolutional DNN transferred from ImageNet, and a Fisher vector classifier on PASCAL VOC 2012 images. Although both models reach similar classification accuracy on the category ``horse'', the authors observed that they use different strategies to classify images of that category. Explanations for a given image are shown in Figure~\ref{figure:validation}(b). The deep neural network looks at the contour of the actual horse, whereas the Fisher vector model (of more rudimentary structure and trained with less data) relies mostly on a copyright tag, that happens to be present on many horse images. Removing the copyright tag in the test images would consequently significantly decrease the measured accuracy of the Fisher vector model but leave the deep neural network predictions unaffected. 

\subsection{Analysis of Scientific Data}
\label{section:scientific}

Beyond model validation, techniques of explanation can also be applied to shed light on scientific problems where human intuition and domain knowledge is often limited. Simple statistical tests and linear models have proved useful to identify correlations between different variables of a system, however, the measured correlations typically remain weak due to the inability of these models to capture the underlying complexity and nonlinearity of the studied problem. For a long time, the computational scientist would face a tradeoff between interpretability and predictive power, where linear models would sometimes be preferred to nonlinear models despite their lower predictive power. We give below a selection of recent works in various fields of research, that combine deep neural networks and explanation techniques to extract insight on the studied scientific problems.

\begin{figure}[!ht]
\centering
\includegraphics[width=1.0\linewidth]{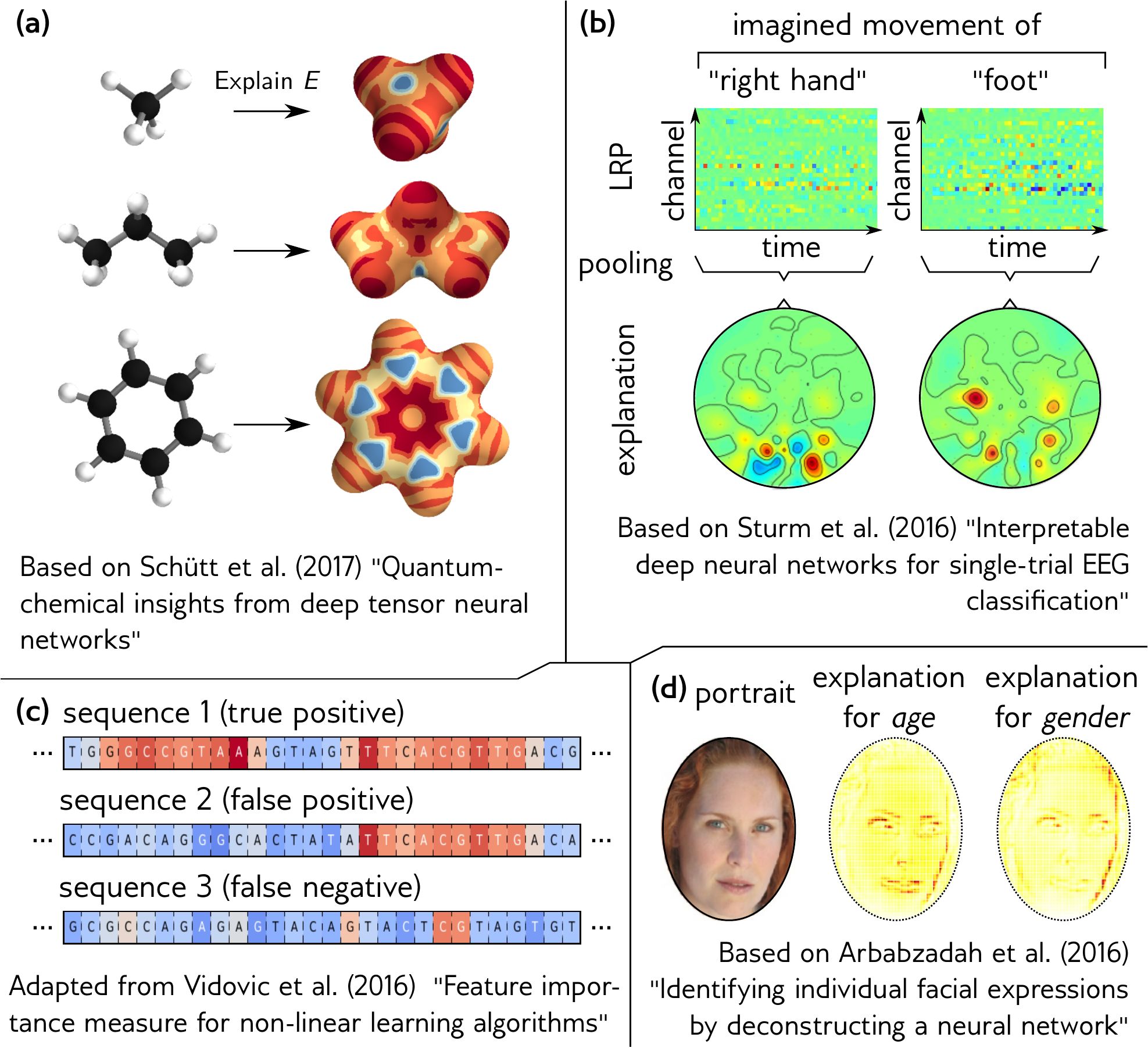}
\caption{Overview of several applications of machine learning explanation techniques in the sciences. (a) Molecular response maps for quantum chemistry, (b) EEG heatmaps for neuroimaging, (c) extracting relevant information from gene sequences, (d) analysis of facial appearance.}
\label{figure:scientific}
\end{figure}

In the domain of atomistic simulations, powerful machine learning models have been produced to link molecular structure to electronic properties \citep{Montavon2013, Hansen2015, Schutt2017,Chmiela2017}. These models  have been trained in a data-driven manner, without simulated physics involved into the prediction. In particular, \citet{Schutt2017} proposed a deep tensor neural network model that incorporates sufficient structure and representational power to simultaneously achieve high predictive power and explainability. Using a test-charge perturbation analysis (a variant of sensitivity analysis where one measures the effect on the neural network output of inserting a charge at a given location), three-dimensional response maps were produced that highlight for each individual molecule spatial structures that were the most relevant for explaining the modeled structure-property relationship. Example of response maps are given in Figure~\ref{figure:scientific}(a) for various molecules.

\citet{Sturm2016} showed that explanation techniques can also be applied to EEG brain recording data. Because the input EEG pattern can take different forms (due to different users, environments, or calibration of the acquisition device), it is important to produce an individual explanation that adapts to these parameters. After training a neural network to map EEG patterns to a set of movements imagined by the user (``right hand'' and ``foot''), a LRP decomposition of that prediction could be achieved in the EEG input domain (a spatiotemporal signal capturing the electrode measurements at various positions on the skull and at multiple time steps), and pooled temporally to produce EEG heatmaps revealing from which part of the brain the decision for ``right hand'' or ``foot'' originates. An interesting property of decomposition techniques in this context is that temporally pooling preserves the total function value, and thus, still corresponds to a decomposition of the prediction. Example of these individual EEG brain maps are given in Figure~\ref{figure:scientific}(b). For classical linear explanation of neural activation patterns in cognitive brain science experiments or Brain Computer Interfacing, see \citep{Blankertz2008, DBLP:journals/neuroimage/LemmBDM11, DBLP:journals/neuroimage/BlankertzLTHM11, DBLP:journals/neuroimage/HaufeMGDHBB14}.

Deep neural networks have also been proposed to make sense of the human genome. \citet{Alipanahi2015} trained a convolutional neural network to map the DNA sequence to protein binding sites. In a second step, they asked what are the nucleotides of that sequence that are the most relevant for explaining the presence of these binding sites. For this, they used a perturbation-based analysis, similar to the sensitivity analysis described in Section \ref{section:explanation-sensitivity}, where the relevance score of each nucleotide is measured based on the effect of mutating it on the neural network prediction. Other measures of feature importance for individual gene sequences have been proposed \citep{DBLP:journals/corr/VidovicGMK16} that apply to a broad class of nonlinear models, from deep networks to weighted degree kernel classifiers. Examples of heatmaps representing relevant genes for various sequences and prediction outcomes are shown in Figure~\ref{figure:scientific}(c).

Explanation techniques also have a potential application in the analysis of face images. These images may reveal a wide range of information about the person's identity, emotional state, or health. However, interpreting them directly in terms of actual features of the input image can be difficult. \citet{DBLP:conf/dagm/ArbabzadahMMS16} applied a LRP technique to identify which pixels in a given image are responsible for explaining, for example, the age and gender attributes. Example of pixel-wise explanations are shown in Figure~\ref{figure:scientific}(d).

\section{Conclusion}

Building transparent machine learning systems is a convergent approach to both extracting novel domain knowledge and performing model validation. As machine learning is increasingly used in real-world decision processes, the necessity for transparent machine learning will continue to grow. Examples that illustrate the limitations of black-box methods were mentioned in Section~\ref{section:validation}.

This tutorial has covered two key directions for improving machine learning transparency: \emph{interpreting} the concepts learned by a model by building prototypes, and \emph{explaining} of the model's decisions by identifying the relevant input variables. The discussion mainly abstracted from the exact choice of deep neural network, training procedure, or application domain. Instead, we have focused on the more conceptual developments, and connected them to recent practical successes reported in the literature.

In particular we have discussed the effect of linking prototypes to the data, via a data density function or a generative model. We have described the crucial difference between sensitivity analysis and decomposition in terms of what these analyses seek to explain. Finally, we have outlined the benefit in terms of robustness, of treating the explanation problem with graph propagation techniques rather than with standard analysis techniques.

This tutorial has focused on post-hoc interpretability, where we do not have full control over the model's structure. Instead, the techniques of interpretation should apply to a general class of nonlinear machine learning models, no matter how they were trained and who trained them  -- even fully trained models that are available for download like BVLC CaffeNet \citep{DBLP:conf/mm/JiaSDKLGGD14} or GoogleNet~\citep{DBLP:conf/cvpr/SzegedyLJSRAEVR15}

In that sense the presented novel technological development in ML allowing for interpretability is an orthogonal strand of research independent of new developments for improving neural network models and their learning algorithms. We would like to stress that all new developments can in this sense always profit in addition from interpretability.

\section*{Acknowledgments}

We gratefully acknowledge discussions and comments on the manuscript by our colleagues Sebastian Lapuschkin, and Alexander Binder. This work was supported by the Brain Korea 21 Plus Program through the National Research Foundation of Korea; the Institute for Information \& Communications Technology Promotion (IITP) grant funded by the Korea government [No.\ 2017-0-00451]; the Deutsche Forschungsgemeinschaft (DFG) [grant MU 987/17-1]; and the German Ministry for Education and Research as Berlin Big Data Center (BBDC) [01IS14013A]. This publication only reflects the authors views. Funding agencies are not liable for any use that may be made of the information contained herein.

\bibliographystyle{elsarticle-harv}

\bibliography{tutorial}

\end{document}